\documentclass[10pt,twocolumn,letterpaper]{article}
\usepackage[pagenumbers]{wacv}
\usepackage{graphicx}
\usepackage{amsmath} 
\usepackage{amssymb}
\usepackage{booktabs}
\usepackage{booktabs}
\usepackage{xcolor}
\usepackage{multirow}
\usepackage{caption}
\usepackage{colortbl} 
\definecolor{lightred}{HTML}{FFF7F7}
\definecolor{lightbule}{HTML}{F2F2FE}
\usepackage{makecell}
\usepackage[pagebackref,breaklinks,colorlinks]{hyperref}
\usepackage[T1]{fontenc}
\usepackage[accsupp]{axessibility}
\definecolor{custompink}{RGB}{251,49,153}

\usepackage[capitalize]{cleveref}
\crefname{section}{Sec.}{Secs.}
\Crefname{section}{Section}{Sections}
\Crefname{table}{Table}{Tables}
\crefname{table}{Tab.}{Tabs.}

\def\confName{WACV}
\def\confYear{2025}

\begin{document}

\title{ReFu: Recursive Fusion for Exemplar-Free 3D Class-Incremental Learning}

\author{
Yi Yang$^{1}$ \quad Lei Zhong$^{1}$ \quad Huiping Zhuang$^{2}$\\
$^{1}$The University of Edinburgh \quad $^{2}$South China University of Technology\\
{\tt\small \{y.yang-249, L.Zhong-10\}@sms.ed.ac.uk \quad hpzhuang@scut.edu.cn}
}
\maketitle

\thispagestyle{empty}
\pagestyle{empty}  

\begin{abstract}
We introduce a novel Recursive Fusion model, dubbed ReFu, designed to integrate point clouds and meshes for exemplar-free 3D Class-Incremental Learning, where the model learns new 3D classes while retaining knowledge of previously learned ones. Unlike existing methods that either rely on storing historical data to mitigate forgetting or focus on single data modalities, ReFu eliminates the need for exemplar storage while utilizing the complementary strengths of both point clouds and meshes. To achieve this, we introduce a recursive method which continuously accumulates knowledge by updating the regularized auto-correlation matrix. Furthermore, we propose a fusion module, featuring a Pointcloud-guided Mesh Attention Layer that learns correlations between the two modalities. This mechanism effectively integrates point cloud and mesh features, leading to more robust and stable continual learning. Experiments across various datasets demonstrate that our proposed framework outperforms existing methods in 3D class-incremental learning. \textbf{Project Page:} \href{https://arlo-yang.github.io/ReFu/}{\textcolor{custompink}{https://arlo-yang.github.io/ReFu/}}

\providecommand{\keywords}[1]
{
  \textbf{\text{Keywords: }} #1
}
\end{abstract}
\section{Introduction}

Class-incremental learning (CIL)~\cite{zhou2024class, wang2024comprehensive, zhou2024continual, masana2022class} is a machine learning paradigm in which models learn new classes incrementally over time without forgetting previously acquired knowledge. However, CIL faces a significant challenge known as catastrophic forgetting~\cite{french1999catastrophic}, where new data can cause a model to lose information from earlier stages.

Building on the principles of CIL, 3DCIL extends this approach to 3D data, which is increasingly important in applications like robotics \cite{lesort2020continual, ayub2022few } and autonomous driving \cite{truong2024conda, yuan2023peer, mirza2022efficient}. Recent replay-based methods~\cite{dong2021i3dol, dong2023inor, zamorski2023continual, chowdhury2022few, fischer2024inemo} in 3DCIL have made progress, but they rely on storing and replaying subsets of previously encountered data as exemplar—a strategy that can be particularly impractical in scenarios with limited storage capacity~\cite{liu2021lifelong}. This challenge is further exacerbated by the complex nature and large size of 3D data, making space limitations even more severe.

\begin{figure}
\centering
\includegraphics[width=1\columnwidth]{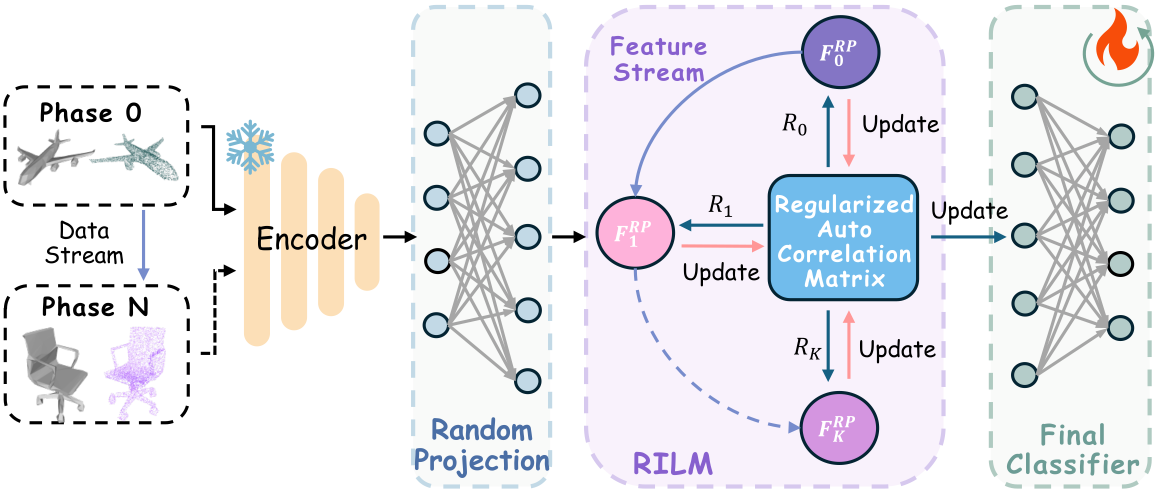} 
\caption{\textbf{Overall scheme of our single-modality Framework:} \textit{RePoint} and \textit{ReMesh}, which consist of a frozen encoder, a random projection layer, the RILM memory module, and a final classifier. RILM facilitates continual learning by recursively updating the regularized auto-correlation matrix, ensuring retention of previously learned categories without forgetting. (Sec.\ref{sec:3.4}, Sec.\ref{sec:3.5}).}
\label{fig1}
\end{figure}

Moreover, existing 3DCIL research mainly focuses on single-modality approaches, typically using point clouds \cite{dong2021i3dol, dong2023inor, zamorski2023continual, chowdhury2022few, liu2021l3doc, chowdhury2021learning}. While point clouds effectively capture the overall shape of objects, they lack inherent structural organization. In contrast, mesh data consists of interconnected vertices and faces, forming continuous surfaces that provide more detailed and coherent structural information than point clouds \cite{liang2022meshmae}. This disparity raises an important question: \textit{Could we leverage the strengths of mesh modalities to assist in 3D point cloud continual learning?}

In this work, we present Recursive Fusion model, dubbed ReFu, a novel framework for exemplar-free 3D Class-Incremental Learning that integrates both point clouds and meshes. At the core of our approach is the Recursive Incremental Learning Mechanism (RILM), which serves as an memory module for 3D data. Rather than storing data exemplars, RILM updates regularized auto-correlation matrices recursively as new data batches are introduced. By maintaining only the updated matrices, RILM overcomes memory constraints and ensures continuous learning without the need for exemplar storage, providing an effective solution for 3D CIL.

Leveraging RILM, we first develop two single-modality frameworks: \textit{RePoint} for point clouds and \textit{ReMesh} for meshes, designed to handle incremental learning tasks independently within their respective modalities, as shown in Fig.~\ref{fig1}. Our experiments demonstrate that both RePoint and ReMesh achieve state-of-the-art performance in their respective 3D incremental learning scenarios. To further enhance performance, we introduce a modality fusion module with a Pointcloud-guided Mesh Attention Layer that adaptively learns correlations between point cloud and mesh features. This attention mechanism computes a spatial map to align the modalities, enabling efficient feature fusion and capturing more comprehensive 3D representations before processing through RILM.

Experiments on the ModelNet40 \cite{wu20153d}, SHREC11 (Pointcloud), and their corresponding mesh datasets Manifold\cite{hu2022subdivision} and SHREC11\cite{lian2011shape} demonstrate that RePoint and ReMesh surpass other baseline models in their respective domains, excelling in both Average Incremental Accuracy and lower Retention Drop across various incremental phases. ReFu, which integrates knowledge from both data modalities, further achieves superior performance. More importantly, unlike previous methods that require storing exemplars \cite{dong2021i3dol, dong2023inor, zamorski2023continual, chowdhury2022few}, ReFu accumulates knowledge by continuously updating regularized auto-correlation matrices without the need for exemplar storage.

\textbf{To summarize, our contributions are:}

\textbf{1)} To our knowledge, ReFu is the first work focus on the multimodal exemplar-free 3DCIL problem using purely 3D data.

\textbf{2)} We introduce a Recursive Incremental Learning Mechanism with a novel fusion module that combines point cloud and mesh features, enabling exemplar-free knowledge accumulation and enhancing 3D representations for robust 3DCIL performance.

\textbf{3)} Experiments demonstrate that ReFu sets a new state of the art in multimodal exemplar-free 3D class-incremental learning, achieving superior results compared to baseline methods.

\section{Related Works}

\subsection{Class-Incremental Learning in 2D:} 
Class-Incremental Learning (CIL) has garnered considerable attention in recent years, with approaches broadly divided into replay-based \cite{rebuffi2017icarl,wang2022foster,rolnick2019experience,bang2021rainbow,liu2023online,liu2021adaptive,prabhu2020gdumb,gopalakrishnan2022knowledge} and exemplar-free methods \cite{kirkpatrick2017overcoming,li2017learning,wang2022learning,wang2022dualprompt,moon2023online,zhou2024revisiting,szatkowski2024adapt}. 

\textbf{Replay-based methods} leverage stored samples or features from previous tasks to reinforce the model's memory while learning new tasks. iCaRL \cite{rebuffi2017icarl} first introduced this approach by selecting samples close to the average embeddings of their respective classes to serve as exemplars. FOSTER \cite{wang2022foster} further utilizes a small memory buffer, introducing a two-stage learning process that first expands the model to handle residuals between new and old categories, followed by compression through distillation. Despite their success, replay-based methods are constrained by privacy \cite{de2021continual} and memory concerns due to the need to store original samples.

\textbf{Exemplar-free methods}, on the other hand, avoid revisiting historical data and are generally categorized into three main streams: regularization-based, prototype-based, and analytic learning-based methods.

\textit{a) Regularization-based methods} prevent catastrophic forgetting by constraining changes to previously learned parameters when learning new tasks. Learning without Forgetting (LwF) \cite{li2017learning} uses knowledge distillation to maintain consistent outputs across old and new tasks, thereby preserving prior knowledge.

\textit{b) Prototype-based methods} improve memory retention by leveraging old class prototypes or generating features that integrate past and present knowledge. Recent approaches \cite{wang2022learning,wang2022dualprompt,moon2023online} employ prompt-based mechanisms to encode both shared and task-specific knowledge, while SimpleCIL \cite{zhou2024revisiting} simplifies the process by using a frozen pretrained model with prototype features of new classes.

\textit{c) Analytic learning-based methods} are a novel category within continual learning, distinguished by their equivalence between continual learning and joint learning frameworks. ACIL \cite{zhuang2022acil} and RanPAC \cite{mcdonnell2024ranpac} tackle CIL through recursive and iterative processes, while GKEAL \cite{zhuang2023gkeal} leverages a Gaussian kernel process to enhance learning in few-shot scenarios, achieving strong performance in data-scarce environments.

\subsection{Class-Incremental Learning in 3D:} 
Compared to 2D class-incremental learning, research on continual learning in the 3D domain has been relatively limited \cite{yang2023geometry,tan2024cross,chowdhury2021learning,liu2021l3doc, chowdhury2022few,zamorski2023continual,dong2023inor,dong2021i3dol}. I3DOL \cite{dong2021i3dol} introduces adaptive geometric structures and geometric-aware attention mechanisms, along with fairness compensation strategies and exemplar support. InOR-Net \cite{dong2023inor} extends this by incorporating category-guided geometric reasoning and critic-induced geometric attention mechanisms, further improving 3D feature recognition. It also uses a dual adaptive fairness compensation strategy to balance performance between new and old tasks. RCR \cite{zamorski2023continual} employs random downsampling and reconstruction loss to compress point cloud data, maintaining memory of prior tasks. \cite{chowdhury2022few} utilize Microshapes to extract shared geometric features, reducing cross-task knowledge loss in few-shot 3D learning. These methods all use small memory buffers to preserve previous task knowledge during incremental learning.

In contrast, earlier works like L3DOC \cite{liu2021l3doc} and \cite{chowdhury2021learning} explored continual learning via knowledge distillation. L3DOC utilizes a shared knowledge base and hierarchical point-knowledge factorization for task-specific knowledge activation, while \cite{chowdhury2021learning} use soft labels from prior models, combined with semantic information, to help the new model retain old class knowledge.

\begin{figure*}
\centering
\includegraphics[width=1\textwidth]{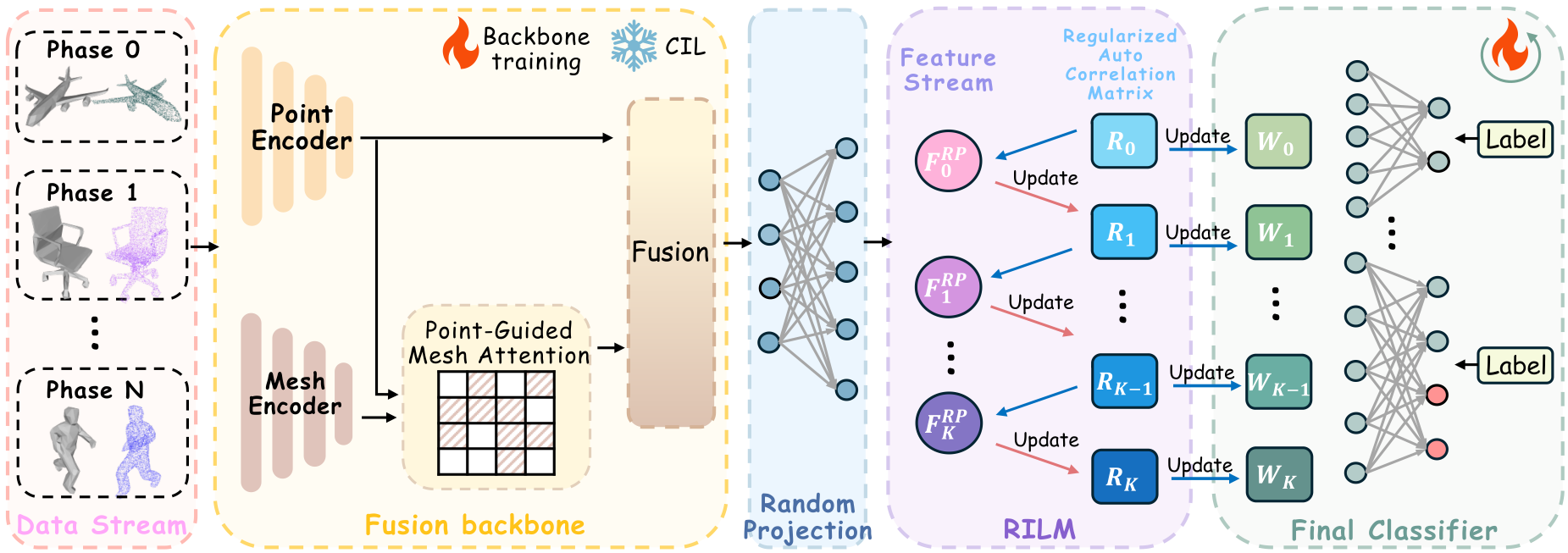} 
\caption{\textbf{Overview of our ReFu framework.} During incremental learning, data flows progressively into the model. As outlined in Section~\ref{sec:3.3}, our proposed fusion backbone, pre-trained on the ShapeNet dataset and frozen during learning, extracts and fuses features from point clouds and meshes. These fused features are then expanded via a random projection layer and input into the Recursive Incremental Learning Mechanism (RILM). RILM recursively updates the regularized auto-correlation matrix and classifier weights. Only the matrix and weights from the previous phase $(n-1)$ are stored, without retaining any raw data as exemplar.}
\label{fig2}
\end{figure*}

\section{Methods}
We start by introducing the necessary preliminaries in Section~\ref{sec:3.1}. Section~\ref{sec:3.2} provides an overview of the proposed model. In Section~\ref{sec:3.3}, we introduce our fusion module, which combines the point-cloud guided mesh attention and the fusion layer for ReFu. Sections~\ref{sec:3.4} and~\ref{sec:3.5} detail the random projection layer and the Recursive Incremental Learning Mechanism (RILM), which are designed to ensure stable memory retention and efficient incremental learning.

\subsection{Preliminaries} \label{sec:3.1}
Class-Incremental Learning (CIL) aims to incrementally learn new classes while retaining knowledge from previous phases. In each phase \(n\), the training set is denoted as \(\mathcal{D}_{n}^{\text{train}} = \left\{ \boldsymbol{X}_{n}^{\text{train}}, \boldsymbol{Y}_{n}^{\text{train}} \right\}\), where the input single modality data \(\boldsymbol{X}_{n}^{\text{train}}\) is either point cloud or mesh, and \(\boldsymbol{Y}_{n}^{\text{train}}\) is the corresponding one-hot encoded label set. For multi-modality, the training set is \(\mathcal{D}_n^{\text{train}} = \left\{ \boldsymbol{X}_{p,n}^{\text{train}}, \boldsymbol{X}_{m,n}^{\text{train}}, \boldsymbol{Y}_n^{\text{train}} \right\}\), where both point cloud \(\boldsymbol{X}_{p,n}^{\text{train}}\) and mesh \(\boldsymbol{X}_{m,n}^{\text{train}}\) data share the same label set \(\boldsymbol{Y}_n^{\text{train}}\).

During each incremental phase, the model is only allowed to access the current phase’s training data. The newly introduced classes are disjoint from previous ones, i.e., \(\boldsymbol{Y}_n \cap \boldsymbol{Y}_{n^{\prime}} = \varnothing\) for \(n \neq n^{\prime}\). After training on the current phase’s data, the model is evaluated on the test set consisting of all seen classes up to phase \(n\), represented as \(\mathcal{Y}_n^{\text{test}} = \boldsymbol{Y}_1^{\text{test}} \cup \dots \cup \boldsymbol{Y}_n^{\text{test}}\).

\subsection{Model Overview} \label{sec:3.2}
Our proposed approach comprises two single-modality frameworks, RePoint for point clouds and ReMesh for meshes (Fig.~\ref{fig1}), alongside a multimodal framework, ReFu, which integrates both modalities to enhance performance in 3D Class-Incremental Learning (3DCIL) (Fig.~\ref{fig2}).

For RePoint and ReMesh, we employ self-supervised pre-trained models, PointMAE \cite{pang2022masked} and MeshMAE \cite{liang2022meshmae}, as encoders. These models are chosen for their label- and class-free pre-training \cite{he2022masked}, making them ideal for incremental learning tasks. Each framework also incorporates a random projection layer for feature expansion and the Recursive Incremental Learning Mechanism (RILM) for efficient knowledge retention.

ReFu integrates both modalities through our fusion module, as discussed in Section~\ref{sec:3.3}. The rest of the process is similar to RePoint and ReMesh, where the encoders are frozen to avoid error accumulation, as suggested by \cite{gan2023decorate}.

\subsection{Fusion} \label{sec:3.3}
Unlike single-modality methods, our ReFu framework emphasizes effective modality integration by first training the fusion backbone on the ShapeNet dataset \cite{chang2015shapenet}. Given a point cloud input \( \boldsymbol{X}_{p}^{\text{train}} \) and a corresponding mesh input \( \boldsymbol{X}_{m}^{\text{train}} \), the point cloud and mesh encoders extract point cloud features \( \boldsymbol{F}_p \in \mathbb{R}^{K \times d} \) and mesh features \( \boldsymbol{F}_m \in \mathbb{R}^{K \times d} \), respectively, where \( N \) represents the number of input samples and \( d \) is the dimensionality of the feature space.

To effectively combine these multimodal features, we employ the Pointcloud-guided Mesh Attention Layer to adaptively learn the correlations between the point cloud and mesh features. We first compute the spatial attention map through the following steps:
\[
\begin{aligned}
& \text{\textit{\textbf{Score}}}_p = \mathit{tanh}\left( \boldsymbol{F}_p \boldsymbol{W}_p \right) \\
& \text{\textit{\textbf{Score}}}_m = \mathit{tanh}\left( \boldsymbol{F}_m \boldsymbol{W}_m \right) \\
& \boldsymbol{W}_{\text{Spa}} = \mathit{Softmax}\left( \text{\textit{\textbf{Score}}}_m \circ \text{\textit{\textbf{Score}}}_p \right)
\end{aligned} \tag{1}
\]
where \( \boldsymbol{W}_p \in \mathbb{R}^{d \times d} \) and \( \boldsymbol{W}_m \in \mathbb{R}^{d \times d} \) are learnable projection matrices. The function \( \mathit{tanh}(\cdot) \) serves as the non-linear activation function, and \( \circ \) denotes the Hadamard product operation. The resulting spatial attention map \( \boldsymbol{W}_{\text{Spa}} \in \mathbb{R}^{d \times d} \) is used to re-weight the mesh features, producing the adjusted representations:
\[
\boldsymbol{F}_m^{\prime} = \boldsymbol{W}_{\text{Spa}}\boldsymbol{F}_m\tag{2}
\]
where \( \boldsymbol{F}_m^{\prime}\) represents the mesh features after spatial attention weighting.

Finally, the processed point cloud and mesh features are concatenated and passed to a classifier to obtain the final prediction:
\[
\hat{y} = \operatorname{CLS}\left( \mathit{concat}\left( \boldsymbol{F}_p, \boldsymbol{F}_m^{\prime} \right) \right),\tag{3}
\]
where \( \mathit{concat}\left( \boldsymbol{F}_p, \boldsymbol{F}_m^{\prime} \right) \) denotes the concatenated feature representation, \( \operatorname{CLS}(\cdot) \) represents the classifier. The output \( \hat{y} \) is the predicted class label.

\subsection{RILM Alignment} \label{sec:3.4}
For the initial phase \( 0 \) in incremental learning, the data \( \boldsymbol{X}_{s,0}^{\text{train}} \) is passed through the encoders to extract the feature matrix. This matrix is then processed by a two-layer linear feed-forward network, where the first layer performs feature expansion via random projection, and the second layer is used for classification. 

Specifically, we feed the input samples \( \boldsymbol{X}_{s,0}^{\text{train}} \) of the classes in phase 0 into the encoder to extract the embeddings for point clouds or meshes, which are then expanded using random projection and processed by an activation function:
\[
\boldsymbol{F}_0^{\text{RP}} = \boldsymbol{\sigma} \left( \mathit{Encoder} \left( \boldsymbol{X}_{s,0}^{\text{train}} \right) \boldsymbol{W}^{\text{RP}} \right)\tag{4}
\]
Here, \( \boldsymbol{\sigma} \) is the activation function. \( \boldsymbol{W}^{\text{RP}} \) denotes the parameters of the random projection layer, and \( \boldsymbol{F}_0^{\text{RP}} \in \mathbb{R}^{N_0 \times d_{\text{RP}}} \), where \( d_{\text{RP}} \) is the expanded dimensionality. The random projection layer expands the feature space, introducing additional parameters and improving memory retention in RILM, leading to enhanced model performance, as analyzed in Section~\ref{sec:4.7}.

Next, we map the expanded embeddings to the label matrix \( \boldsymbol{Y}_0^{\text{train}} \) via a linear classifier layer, whose weights are computed by solving the following:
\[
\underset{\boldsymbol{W}_0}{\operatorname{argmin}} = \left\|\boldsymbol{Y}_0^{\text{train}} - \boldsymbol{F}_0^{\text{RP}} \boldsymbol{W}_0 \right\|_F^2 + \eta \left\|\boldsymbol{W}_0\right\|_F^2\tag{5}
\]
where \( \|\cdot\|_F \) denotes the Frobenius norm, and \( \eta \) is the regularization term. The optimal solution is then obtained as:
\[
\hat{\boldsymbol{W}}_0 = \left(\left(\boldsymbol{F}_0^{\text{RP}}\right)^\top \boldsymbol{F}_0^{\text{RP}} + \eta \boldsymbol{I}\right)^{-1} \left(\boldsymbol{F}_0^{\text{RP}}\right)^\top \boldsymbol{Y}_0^{\text{train}}\tag{6}
\] 
where \( \hat{\boldsymbol{W}}_0 \) represents the estimated parameters of the linear classifier layer, and \( { }^\top \) denotes the transpose operation.

\subsection{Class-Incremental Learning} \label{sec:3.5}
Following the RILM alignment of point cloud and mesh embeddings, we proceed to the class-incremental learning (CIL) steps. Specifically, the learning problem using all seen data up to phase \( n-1 \) can be extended from (5) as:
\[
\underset{\boldsymbol{W}_{n-1}}{\operatorname{argmin}} \left\|\boldsymbol{Y}_{0:n-1}^{\text{train}} - \boldsymbol{F}_{0:n-1}^{\text{RP}} \boldsymbol{W}_{n-1} \right\|_F^2 + \eta \left\|\boldsymbol{W}_{n-1}\right\|_F^2 \tag{7}
\]
where
\small
\[
\boldsymbol{Y}_{0:n-1}^{\text{train}} = \left[\begin{array}{ccc}
\boldsymbol{Y}_0^{\text{train}} & \cdots & \mathbf{0} \\
\vdots & \ddots & \vdots \\
\mathbf{0} & \cdots & \boldsymbol{Y}_{n-1}^{\text{train}}
\end{array}\right], \quad
\boldsymbol{F}_{0:n-1}^{\text{RP}} = \left[\begin{array}{c}
\boldsymbol{F}_0^{\text{RP}} \\
\vdots \\
\boldsymbol{F}_{n-1}^{\text{RP}}
\end{array}\right] \tag{8}
\]
\normalsize

Similar to (6), the solution to this optimization problem can be obtained as:
\[
\hat{\boldsymbol{W}}_{n-1} = \left(\sum_{n'=0}^{n-1} \boldsymbol{A}_{n'} + \eta \boldsymbol{I}\right)^{-1} \left(\sum_{n'=0}^{n-1} \boldsymbol{C}_{n'}\right) \tag{9}
\]
\noindent where
\begin{equation}
\left\{
\begin{aligned}
\boldsymbol{A}_{n'} & = \left(\boldsymbol{F}_{n'}^{\mathrm{RP}}\right)^{\top} \boldsymbol{F}_{n'}^{\mathrm{RP}} \\
\boldsymbol{C}_{n'} & = \left(\boldsymbol{F}_{n'}^{\mathrm{RP}}\right)^{\top} \boldsymbol{Y}_{n'}^{\mathrm{train}}
\end{aligned}
\right.
\tag{10}
\end{equation}

\noindent Here, \( \boldsymbol{A}_{n'} \) represents the auto-correlation feature matrix, and \( \boldsymbol{C}_{n'} \) represents the cross-correlation matrix. To further simplify the equation, let:
\[
\boldsymbol{R}_{n-1} = \left(\sum_{n'=0}^{n-1} \boldsymbol{A}_{n'} + \eta \boldsymbol{I}\right)^{-1} \tag{11}
\]
\noindent to be the regularized auto-correlation matrix. We can redefine the CIL process as a recursive least squares task as outlined in the following theorem.

\

\noindent \textbf{Theorem 1}. Given training data \( \mathcal{D}_n^{\text{train}} \) and the estimated weights of the final classifier layer \( \hat{\boldsymbol{W}}_{n-1} \) from phase \( n-1 \), the updated weights \( \hat{\boldsymbol{W}}_n \) can be recursively obtained by:
\[
\hat{\boldsymbol{W}}_n = \hat{\boldsymbol{W}}_{n-1} - \boldsymbol{R}_n \boldsymbol{A}_n \hat{\boldsymbol{W}}_{n-1} + \boldsymbol{R}_n \boldsymbol{C}_n \tag{12}
\]
\noindent with
\small
\begin{align*}
\boldsymbol{R}_n &= \boldsymbol{R}_{n-1} - \boldsymbol{R}_{n-1}\left(\boldsymbol{F}_n^{\mathrm{RP}}\right)^\top \\
&\quad \times \left(\boldsymbol{F}_n^{\mathrm{RP}} \boldsymbol{R}_{n-1} \left(\boldsymbol{F}_n^{\mathrm{RP}}\right)^\top + \boldsymbol{I}\right)^{-1}  \boldsymbol{F}_n^{\mathrm{RP}} \boldsymbol{R}_{n-1} \tag{13}
\end{align*}
\normalsize

\noindent \textit{Proof}. See Supplementary Material for details.\vspace{1\baselineskip}

\noindent \textbf{Summary:} Theorem 1 proves that joint training can be transformed into a recursive incremental learning process. Instead of requiring access to the entire dataset at once, we only need to retain the weights \( \hat{\boldsymbol{W}}_{n-1} \) and the regularized auto-correlation matrix \( \boldsymbol{R}_{n-1} \) from the previous phase. By recursively updating these with new data, while keeping the fusion backbone frozen, the class-incremental learning (CIL) process becomes mathematically equivalent to joint training. This demonstrates the strong memory retention capability of our RILM.

\begingroup
\setlength{\abovecaptionskip}{0pt}  
\setlength{\belowcaptionskip}{-6pt}  
\begin{table*}[ht]
    \captionsetup{justification=raggedright,singlelinecheck=false}
    \caption{\textbf{Performance comparison on point cloud datasets}, evaluated using \(\mathcal{A}\) and \(\mathcal{R}\). Bold values denote the overall best results, while underlined values highlight the top-performing baselines. Red-highlighted rows indicate our RePoint method, and blue-highlighted rows represent ReFu. ReFu is separated from other methods as it leverages both point cloud and mesh inputs, provided here only for direct comparison (see Section~\ref{sec:4.6}).}
    \label{table1}
    \renewcommand{\arraystretch}{1.3} 
    \begin{center}
    \scalebox{0.85}{
    \begin{tabular}{lcccccccccc}
        \toprule
        \multicolumn{2}{c}{} & \multicolumn{4}{c}{\textbf{SHREC11 (Point cloud)}} & \multicolumn{4}{c}{\textbf{ModelNet40}} \\
        \cmidrule(lr){3-6} \cmidrule(lr){7-10}
        \textbf{Method} & \textbf{Exemplar-free?} & \multicolumn{2}{c}{\textbf{10 Phase}} & \multicolumn{2}{c}{\textbf{30 Phase}} & \multicolumn{2}{c}{\textbf{10 Phase}} & \multicolumn{2}{c}{\textbf{40 Phase}} \\
        \cmidrule(lr){3-4} \cmidrule(lr){5-6} \cmidrule(lr){7-8} \cmidrule(lr){9-10}
        & & \(\mathcal{A}\) (\%) $\uparrow$ & \(\mathcal{R}\) (\%) $\downarrow$ & \(\mathcal{A}\) (\%) $\uparrow$ & \(\mathcal{R}\) (\%) $\downarrow$ & \(\mathcal{A}\) (\%) $\uparrow$ & \(\mathcal{R}\) (\%) $\downarrow$ & \(\mathcal{A}\) (\%) $\uparrow$ & \(\mathcal{R}\) (\%) $\downarrow$ \\
        \midrule
        Fine-tuning & $\checkmark$ & 30.39 & 87.34 & 13.32 & 96.67 & 28.26 & 89.62 & 9.70 & 99.18 \\
        LwF \cite{li2017learning} & $\checkmark$ & 42.21 & 81.27 & 12.32 & 94.52 & 48.80 & 75.52 & 11.59 & 95.94 \\
        SimpleCIL \cite{zhou2024revisiting}  & $\checkmark$ & 76.56 & 28.37 & 77.11 & 31.15 & 79.77 & 25.65 &  81.28 & 33.06 \\
        iCaRL \cite{rebuffi2017icarl}  & $\times$ & \underline{89.56} & 17.02 & 81.73 & 27.70 & 87.01 & 27.22 & 82.18 & 36.71 \\
        Foster \cite{wang2022foster}  & $\times$ & 88.56 &  \underline{14.92} & \underline{92.36} & \underline{14.33} & \underline{92.74} & \underline{14.33} & \underline{92.70} & \underline{14.38} \\
        \rowcolor{lightred}
        RePoint (Ours) & $\checkmark$ & \textbf{94.82} & \textbf{8.73} & \textbf{94.61} & \textbf{9.29} & \textbf{96.51} & \textbf{7.65} & \textbf{96.85} & \textbf{8.20} \\
        \midrule
        \rowcolor{lightbule}
        ReFu (Ours) & $\checkmark$ & \textbf{96.40} & \textbf{6.73} & \textbf{96.36} & \textbf{6.93} & \textbf{97.42} & \textbf{5.12} & \textbf{97.21} & \textbf{6.51} \\
        \bottomrule
    \end{tabular}}
    \end{center}
\end{table*}
\endgroup
\begingroup
\setlength{\abovecaptionskip}{-0pt}  
\setlength{\belowcaptionskip}{-6pt}  
\begin{table*}[ht]
    \captionsetup{justification=raggedright,singlelinecheck=false}
     \caption{\textbf{Performance comparison on mesh datasets}, evaluated using \(\mathcal{A}\) and \(\mathcal{R}\). Bold values denote the overall best results, while underlined values highlight the top-performing baselines. Red-highlighted rows indicate our ReMesh method, and blue-highlighted rows represent ReFu. ReFu is separated from other methods as it leverages both point cloud and mesh inputs, provided here only for direct comparison (see Section~\ref{sec:4.6}).}
    \label{table2}
    \renewcommand{\arraystretch}{1.3} 
    \begin{center}
    \scalebox{0.85}{
    \begin{tabular}{lcccccccccc}
        \toprule
        \multicolumn{2}{c}{} & \multicolumn{4}{c}{\textbf{SHREC11 (Mesh)}} & \multicolumn{4}{c}{\textbf{Manifold40}} \\
        \cmidrule(lr){3-6} \cmidrule(lr){7-10}
        \textbf{Method} & \textbf{Exemplar-free?} & \multicolumn{2}{c}{\textbf{10 Phase}} & \multicolumn{2}{c}{\textbf{30 Phase}} & \multicolumn{2}{c}{\textbf{10 Phase}} & \multicolumn{2}{c}{\textbf{40 Phase}} \\
        \cmidrule(lr){3-4} \cmidrule(lr){5-6} \cmidrule(lr){7-8} \cmidrule(lr){9-10}
        & & \(\mathcal{A}\) (\%) $\uparrow$ & \(\mathcal{R}\) (\%) $\downarrow$ & \(\mathcal{A}\) (\%) $\uparrow$ & \(\mathcal{R}\) (\%) $\downarrow$ & \(\mathcal{A}\) (\%) $\uparrow$ & \(\mathcal{R}\) (\%) $\downarrow$ & \(\mathcal{A}\) (\%) $\uparrow$ & \(\mathcal{R}\) (\%) $\downarrow$ \\
        \midrule
        Fine-tuning & $\checkmark$ & 28.65 & 90.00 & 13.31 & 96.67 & 26.86 & 88.61 & 9.70 & 99.19 \\
        LwF \cite{li2017learning}  & $\checkmark$ & 34.68 & 86.67 & 13.31 & 96.67 & 36.13 & 87.57 & 14.35 & 95.95 \\
        SimpleCIL \cite{zhou2024revisiting}  & $\checkmark$ & 86.29 & 18.77 & 86.87 & 18.77 & 77.48 & 27.91 & 79.26 & 36.06 \\
        iCaRL \cite{rebuffi2017icarl}  & $\times$ & 85.20 & 21.42 & 84.22 & 25.79 & 76.14 & 38.03 & 73.79 & 47.11 \\
        Foster \cite{wang2022foster}  & $\times$ & \underline{92.94} & \underline{13.73} & \underline{91.83} &  \underline{13.41} & \underline{86.10} & \underline{22.81} & \underline{83.66} & \underline{25.28} \\
        \rowcolor{lightred}
        ReMesh (Ours) & $\checkmark$ & \textbf{95.19} & \textbf{7.82} & \textbf{95.54} & \textbf{8.21} & \textbf{94.08} & \textbf{11.31} &\textbf{94.51} & \textbf{12.61} \\
        \midrule
        \rowcolor{lightbule}
        ReFu (Ours) & $\checkmark$ & \textbf{96.40} & \textbf{6.73} & \textbf{96.36} & \textbf{6.93} & \textbf{97.42} & \textbf{5.12} & \textbf{97.21} & \textbf{6.51} \\
        \bottomrule
    \end{tabular}}
    \end{center}
\end{table*}
\endgroup

\section{Experiments}
\subsection{Datasets}\label{sec:4.1}
We employ the ModelNet40 \cite{wu20153d} dataset as our primary source for rigid object classification and the SHREC11 \cite{lian2011shape} dataset to evaluate performance on non-rigid 3D objects. ModelNet40, which has similar structure to the ShapeNet \cite{chang2015shapenet} dataset used for pretraining, contains 12,311 3D models, with 9,843 used for training and 2,468 for testing, spanning 40 distinct categories. Point cloud representations were generated by uniformly sampling 8,192 points from each object. For mesh data, due to the presence of holes or boundaries on the surface of the original ModelNet40 data, we follow the data pre-processing protocol of MeshMAE \cite{liang2022meshmae} and SubdivNet \cite{hu2022subdivision}, generating remeshed watertight or 2-manifold meshes, referred to as the Manifold40 dataset. No data augmentation techniques were applied to either the ModelNet40 or Manifold40 datasets.

In contrast, the SHREC11 dataset comprises 30 classes with 20 samples per class, focusing primarily on non-rigid 3D objects. These objects exhibit greater morphological flexibility, with their shapes undergoing significant changes under different poses and motions. This dataset was selected to evaluate the robustness of our method when applied to out-of-domain data. For point clouds, we uniformly sampled 8,192 points per object, while for meshes, we adopted the pre-processing protocol from SubdivNet \cite{hu2022subdivision}.

\subsection{Class-Incremental Setting}\label{sec:4.2}
We evaluate class-incremental learning under two settings: a short-range setup with 10 phases and a long-range setup with \( N \) phases, where \( N \) represents the total number of classes. For ModelNet40 (Manifold40), \( N = 40 \), and for SHREC11, \( N = 30 \). In the short-range setup, each phase introduces 4 classes for ModelNet40 and 3 classes for SHREC11, while the long-range setup adds 1 class per phase for both datasets.

\begin{figure*}
\centering
\includegraphics[width=1\textwidth]{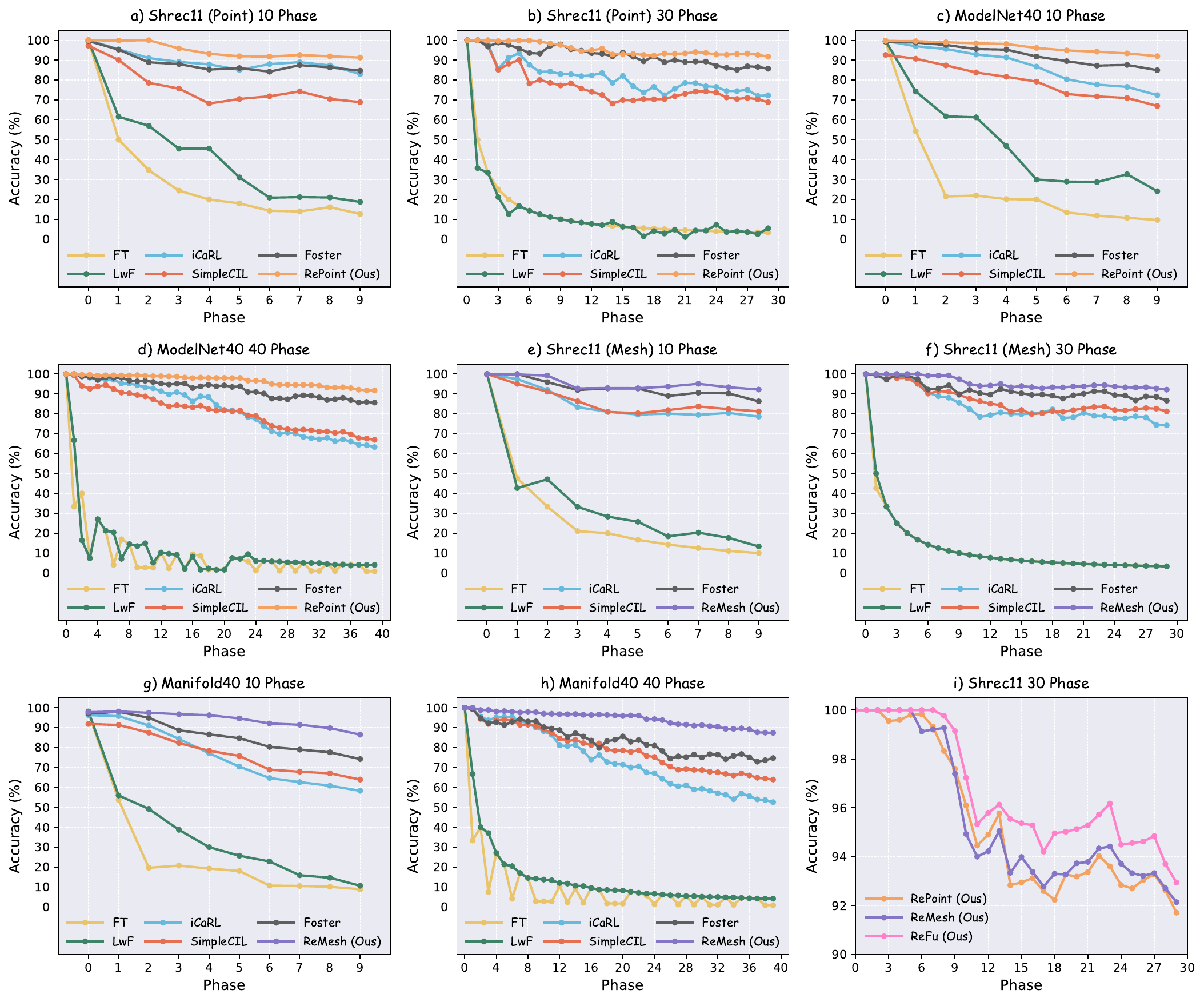} 
\caption{\textbf{(a) $\sim$ (h): }Test accuracy \( \mathcal{A}_n \) at each incremental stage, ranging from \( \mathcal{A}_1^0 \) to \( \mathcal{A}_N^0 \). Here, \( n \) represents the current phase and \( N \) is the total number of phases. The results are based on the PointMAE / MeshMAE backbone. We report accuracy on point cloud datasets: (1) SHREC11 (Point) and (2) ModelNet40, and mesh datasets: (3) SHREC11 and (4) Manifold40, under 10 phase and \(N\) phase settings. \textbf{(i)} shows the test accuracy of our proposed three frameworks, ReFu, RePoint and Remesh.}

\label{fig3}
\end{figure*}

\subsection{Evaluation Metrics}\label{sec:4.3}
Evaluation metrics are designed to assess two critical aspects: Incremental Learning Capability and Knowledge Retention. To evaluate these dimensions, we adopt metrics similar to those used in \cite{yue2024mmal}, focusing on Average Incremental Accuracy (\(\mathcal{A}\)) and Retention Drop (\(\mathcal{R}\)).

For assessing Incremental Learning Capability, we employ the Average Incremental Accuracy (\(\mathcal{A}\)) metric. \(\mathcal{A}\) provides a comprehensive measure of a model's performance across all learning phases and is calculated as \(\mathcal{A} = \frac{1}{N} \sum_{n=1}^N \mathcal{A}_n\), where \( \mathcal{A}_n \) represents the average test accuracy at phase \( n \) across all previously seen classes. A higher \(\mathcal{A}\) value indicates the model's effectiveness in maintaining high performance as it incrementally learns new classes.

To evaluate Knowledge Retention, we introduce the Retention Drop (\(\mathcal{R}\)) metric. \(\mathcal{R}\) quantifies performance degradation as the model learns new tasks and is defined as \(\mathcal{R} = \mathcal{A}_1 - \mathcal{A}_N\), where \( \mathcal{A}_1 \) is the test accuracy after the initial phase, and \( \mathcal{A}_N \) is the accuracy after the final phase \( N \). This metric measures the model's ability to retain knowledge throughout incremental learning. A lower \(\mathcal{R}\) value indicates better retention, reflecting less degradation on early learned tasks as new tasks are introduced.

\subsection{Implementation details}\label{sec:4.4}
We conduct all experiments using PyTorch \cite{paszke2019pytorch}. Following the approach in \cite{chowdhury2022few, tan2024cross}, we replace the encoder in 2DCIL methods with pre-trained PointMAE and MeshMAE models, using them as feature extractors for point cloud and mesh datasets, respectively. For the training of ReFu's fusion backbone, we freeze the self-supervised pre-trained point cloud and mesh encoders and fine-tune the remaining components of the model (pointcloud-guided mesh attention layer, fusion layer, and classifier). We use the Adam optimizer \cite{kingma2014adam} with a learning rate of 4e-3 and a batch size of 24. After this training phase, the parameters of the backbone are frozen, serving as feature extractors in the subsequent incremental learning steps.

\subsection{Performance of \textit{RePoint} and \textit{ReMesh}}\label{sec:4.5}
To assess the efficacy of the newly developed RePoint and ReMesh frameworks, we conducted comparative analyses against various incremental learning methodologies: 1) Fine-tuning, where the model is initialized with the parameters from the previous phase and re-trained on new data batches; 2) LwF \cite{li2017learning}, which utilizes the preservation of outputs from previous examples as a regularizer for new tasks; 3) iCaRL \cite{rebuffi2017icarl}, a method that combines exemplars with distillation to prevent forgetting; 4) SimpleCIL\cite{zhou2024revisiting}, using a fixed feature extractor from a pretrained model and prototypes of each class as classifier weights; 5) Foster\cite{wang2022foster}, which enhances the model’s adaptability to new categories by integrating new trainable feature extractors with the existing model and maintains a portion of old data to preserve the memory of previous categories. 

We compare our proposed RePoint and ReMesh methods with the baseline approaches in Tables~\ref{table1} and \ref{table2}. The results show that both RePoint and ReMesh consistently outperform the compared methods, demonstrating strong performance in 3D class-incremental learning.

For RePoint, on the SHREC11 dataset, our method increases \(\mathcal{A}\) by \(5.26\%\) and decreases \(\mathcal{R}\) by \(6.19\%\) in the 10-phase setting. In the 30-phase setting, it improves \(\mathcal{A}\) by \(2.25\%\) and reduces \(\mathcal{R}\) by \(5.04\%\). On ModelNet40, RePoint surpasses Foster by \(3.77\%\) in \(\mathcal{A}\) and reduces \(\mathcal{R}\) by \(6.68\%\) in the 10-phase scenario, with gains of \(4.15\%\) in \(\mathcal{A}\) and a reduction of \(6.18\%\) in \(\mathcal{R}\) in the 40-phase setting.

For ReMesh, on SHREC11, it improves \(\mathcal{A}\) by \(2.25\%\) and decreases \(\mathcal{R}\) by \(5.91\%\) in the 10-phase setting, and by \(3.71\%\) in \(\mathcal{A}\) and \(5.2\%\) in \(\mathcal{R}\) in the 30-phase setting. On Manifold40, ReMesh outperforms Foster by \(7.98\%\) in \(\mathcal{A}\) and decreases \(\mathcal{R}\) by \(11.5\%\) in the 10-phase scenario, with gains of \(10.85\%\) in \(\mathcal{A}\) and reductions of \(12.67\%\) in \(\mathcal{R}\) in the 40-phase setting.

Furthermore, we present the testing accuracy at each incremental step for RePoint, ReMesh, and other baselines in Fig.~\ref{fig3}. Our methods consistently achieve top performance in most steps in nearly all incremental steps and outperform all the baselines in the final phases.  Similar to the exemplar-free SimpleCIL, RePoint and ReMesh also freeze the backbone during incremental learning, but use a more effective memory mechanism instead of relying on prototypes. In summary, our methods demonstrate clear advantages in 3DCIL, validating their effectiveness.

\subsection{Performance of \textit{ReFu}}\label{sec:4.6}
The fusion of 3D modalities in ReFu further enhances recognition accuracy and knowledge retention in CIL tasks. In Tables~\ref{table1} and \ref{table2}, we also provide the performance of ReFu for comparison with RePoint and ReMesh. The results indicate that ReFu outperforms the single-modality methods, demonstrating the effectiveness of multimodal fusion. Furthermore, in Fig.~\ref{fig3} (i), we present the testing accuracy at each incremental step for ReFu, RePoint, and ReMesh on the SHREC11 30-phase task. It can be observed that ReFu consistently achieves better performance than the single-modality methods at all stages.

\subsection{Ablation Studies}\label{sec:4.7}
\noindent\textbf{Random Projection:}  
As presented in Table~\ref{table3}, Random Projection (RP) is a key component of our model, with its dimensionality governed by the parameter $d_{\text{(rp)}}$. By increasing the feature dimension, RP effectively captures additional information and alleviates the over-fitting issues typically encountered in recursive learning approaches \cite{zhuang2022acil}. To evaluate its impact, we performed a 10-phase incremental learning experiment using ReFu on the SHREC11 and ModelNet40 datasets, where the feature dimensionality was expanded by 12 times compared to the original.\vspace{1\baselineskip}

\noindent\textbf{Fusion Methods:}
We first assess the effectiveness of our fusion strategy by comparing it to a variant utilizing simple addition fusion. As shown in Table~\ref{table4}, addition fusion yields sub-optimal performance. On the SHREC11 dataset, the average accuracy (\(\mathcal{A}\)) is 95.23\%, only marginally outperforming the single-modality ReMesh baseline (95.19\%). On ModelNet40, \(\mathcal{A}\) drops below that of RePoint (96.51\%), indicating that addition fusion struggles to leverage complementary information between modalities. When one modality under-performs, the overall result is adversely impacted, limiting the potential benefits of multimodal fusion.

In contrast, concatenation fusion delivers superior results. On SHREC11, it improves \(\mathcal{A}\) by 1.09\% over addition fusion and reduces retention drop (\(\mathcal{R}\)) by 1.02\%. On ModelNet40, it further increases \(\mathcal{A}\) by 1.59\% and reduces \(\mathcal{R}\) by 1.61\%. These findings suggest that concatenation fusion preserves more information from both modalities, mitigating knowledge forgetting in continual learning.

Nevertheless, simple concatenation does not truly "fuse" the features from different modalities. To address this, we employ attention-guided concatenation fusion. On SHREC11, attention-guided concatenation boosts \(\mathcal{A}\) by an additional 0.08\% over standard concatenation and reduces \(\mathcal{R}\) by 0.52\%. On ModelNet40, it further raises \(\mathcal{A}\) by 0.28\% and decreases \(\mathcal{R}\) by 1.01\%. These results demonstrate that attention-guided fusion effectively enhances cross-modal interaction, leading to improved performance in 3D Class-Incremental Learning (3DCIL).

\begin{table}[t]
    
    \captionsetup{justification=raggedright,singlelinecheck=false}
    \caption{Performance of ReFu methods over 10 phases on SHREC11 and ModelNet40 (Manifold40) datasets, evaluated with \(\mathcal{A}\) and \(\mathcal{R}\). Red-highlighted rows indicate results with Random Projection (RP) applied.}
    \renewcommand{\arraystretch}{1.1}
    \label{table3}
    \scalebox{0.95}{
    \resizebox{\columnwidth}{!}{
    \begin{tabular}{lcccc}
        \toprule
        Dataset & \textbf{With RP?} & \(\mathcal{A}\) (\%) $\uparrow$ & \(\mathcal{R}\) (\%) $\downarrow$ \\
        \midrule
        \multirow{2}{*}{ SHREC11} & $\times$ & 94.13 & 10.82 \\
                                 & \cellcolor{lightred}$\checkmark$ & \cellcolor{lightred}96.40 \textcolor{blue}{(+2.27)} & \cellcolor{lightred}6.73 \textcolor{blue}{(-4.09)} \\
        \midrule
        \multirow{2}{*}{\makecell{ModelNet40 \\ (Manifold40)}} & $\times$ & 94.45 & 12.01 \\
                                              & \cellcolor{lightred}$\checkmark$ & \cellcolor{lightred}97.42 \textcolor{blue}{(+2.97)} & \cellcolor{lightred}5.12 \textcolor{blue}{(-6.89)} \\
        \bottomrule
    \end{tabular}}}
\end{table}

\begin{table}[t]
    \centering
    \captionsetup{justification=raggedright,singlelinecheck=false}
    \caption{Performance of ReFu methods during 10 phases. Metrics used are \(\mathcal{A}\) and \(\mathcal{R}\). For clarity, fusion methods are abbreviated as follows: AddF = addition, ConcatF = concatenation, AttnF = attention-guided concatenation.}
    \renewcommand{\arraystretch}{1.1} 
    \label{table4}
    \scalebox{0.9}{ 
    \resizebox{\columnwidth}{!}{
    \begin{tabular}{lcccc}
        \toprule
        Dataset & \textbf{Fusion Methods} & \(\mathcal{A}\) (\%) $\uparrow$ & \(\mathcal{R}\) (\%) $\downarrow$ \\
        \midrule
        \multirow{3}{*}{ SHREC11} & AddF & 95.23 & 8.27 \\
        & ConcatF & 96.32 \textcolor{blue}{(+1.09)} & 7.25 \textcolor{blue}{(-1.02)} \\
        & \cellcolor{lightred}AttnF & \cellcolor{lightred}96.40 \textcolor{blue}{(+0.08)} & \cellcolor{lightred}6.73 \textcolor{blue}{(-0.52)} \\
        \midrule
        \multirow{3}{*}{\makecell{ModelNet40 \\ (Manifold40)}} & AddF & 95.55 & 7.74 \\
        & ConcatF & 97.14 \textcolor{blue}{(+1.59)} & 6.13 \textcolor{blue}{(-1.61)} \\
        & \cellcolor{lightred}AttnF & \cellcolor{lightred}97.42 \textcolor{blue}{(+0.28)} & \cellcolor{lightred}5.12 \textcolor{blue}{(-1.01)} \\
        \bottomrule
    \end{tabular}}}
\end{table}

\section{Discussion}
\noindent \textbf{Why we use 2D CIL methods as baselines?   } 
Currently, there are very few 3DCIL approaches \cite{fischer2024inemo} designed for meshes, with most research focused on point clouds. For example, methods like RCR \cite{zamorski2023continual} have show that point cloud data, even after significant compression, can still serve as effective exemplars.  However, these methods are not directly applicable to meshes, and there is no evidence suggesting they would achieve comparable performance, limiting direct comparisons. In addition, given the extensive research in 2DCIL \cite{gao2024consistent,ye2024continual,tran2024text, roy2024convolutional}, most 3DCIL approaches \cite{liu2021l3doc,chowdhury2021learning, tan2024cross,dong2023inor, dong2021i3dol} adopt 2D strategies by replacing 2D encoders with 3D counterparts, thereby facilitating the transfer of 2D CIL techniques to 3D tasks. \vspace{1\baselineskip}

\noindent\textbf{The Role of Large Pretrained Models in 3DCIL:   }
Current 3DCIL methods heavily rely on early feature extractors like PointNet \cite{qi2017pointnet}, PointNet++ \cite{qi2017pointnet++}, and DGCNN \cite{wang2019dynamic}. However, leveraging more powerful backbone networks is crucial for improving the performance of 3D continual learning. Studies like \cite{tan2024cross} show that using PointCLIP \cite{zhang2022pointclip} can further improve results. In 2DCIL, SimpleCIL \cite{zhou2024revisiting} demonstrated that frozen pretrained models (PTMs) can even outperform advanced CIL methods by simply updating the classifier with prototype features. PTMs, trained on large datasets, transfer knowledge efficiently and generalize well, making pretrained models a new paradigm in continual learning \cite{liu2023tail,park2024pre,pian2023audio,leestella,lee2023pre}. Motivated by this, we introduce the MAE \cite{he2022masked} pretraining paradigm into 3DCIL, enhancing knowledge transfer and performance.

\section{Conclusion}

In this paper, we introduce the Recursive Fusion model (ReFu), a novel framework for exemplar-free 3D Class-Incremental Learning that integrates both point cloud and mesh modalities. Our model employs the Recursive Incremental Learning Mechanism (RILM) to accumulate knowledge without storing exemplars, recursively updating regularized auto-correlation matrices. We develop two specialized models: \textit{RePoint} for point clouds and \textit{ReMesh} for meshes, both achieving state-of-the-art performance in 3D incremental learning tasks. ReFu further enhances 3D representation learning with the Pointcloud-guided Mesh Attention Layer. Extensive experiments on multiple datasets validate the effectiveness of our approach.

{\small
\bibliographystyle{ieee_fullname}
\bibliography{egbib}

\begin{thebibliography}{10}\itemsep=-1pt

\bibitem{ayub2022few}
Ali Ayub and Carter Fendley.
\newblock Few-shot continual active learning by a robot.
\newblock {\em NIPS}, 2022.

\bibitem{bang2021rainbow}
Jihwan Bang, Heesu Kim, YoungJoon Yoo, Jung-Woo Ha, and Jonghyun Choi.
\newblock Rainbow memory: Continual learning with a memory of diverse samples.
\newblock In {\em CVPR}, 2021.

\bibitem{chang2015shapenet}
Angel~X Chang, Thomas Funkhouser, Leonidas Guibas, Pat Hanrahan, Qixing Huang, Zimo Li, Silvio Savarese, Manolis Savva, Shuran Song, Hao Su, et~al.
\newblock Shapenet: An information-rich 3d model repository.
\newblock {\em arXiv preprint arXiv:1512.03012}, 2015.

\bibitem{chowdhury2022few}
Townim Chowdhury, Ali Cheraghian, Sameera Ramasinghe, Sahar Ahmadi, Morteza Saberi, and Shafin Rahman.
\newblock Few-shot class-incremental learning for 3d point cloud objects.
\newblock In {\em ECCV}, 2022.

\bibitem{chowdhury2021learning}
Townim Chowdhury, Mahira Jalisha, Ali Cheraghian, and Shafin Rahman.
\newblock Learning without forgetting for 3d point cloud objects.
\newblock In {\em Advances in Computational Intelligence: 16th International Work-Conference on Artificial Neural Networks, IWANN}, 2021.

\bibitem{de2021continual}
Matthias De~Lange, Rahaf Aljundi, Marc Masana, Sarah Parisot, Xu Jia, Ale{\v{s}} Leonardis, Gregory Slabaugh, and Tinne Tuytelaars.
\newblock A continual learning survey: Defying forgetting in classification tasks.
\newblock {\em PAMI}, 2021.

\bibitem{dong2021i3dol}
Jiahua Dong, Yang Cong, Gan Sun, Bingtao Ma, and Lichen Wang.
\newblock I3dol: Incremental 3d object learning without catastrophic forgetting.
\newblock In {\em AAAI}, 2021.

\bibitem{dong2023inor}
Jiahua Dong, Yang Cong, Gan Sun, Lixu Wang, Lingjuan Lyu, Jun Li, and Ender Konukoglu.
\newblock Inor-net: Incremental 3-d object recognition network for point cloud representation.
\newblock {\em IEEE Transactions on Neural Networks and Learning Systems}, 2023.

\bibitem{fischer2024inemo}
Tom Fischer, Yaoyao Liu, Artur Jesslen, Noor Ahmed, Prakhar Kaushik, Angtian Wang, Alan Yuille, Adam Kortylewski, and Eddy Ilg.
\newblock inemo: Incremental neural mesh models for robust class-incremental learning.
\newblock {\em ECCV}, 2024.

\bibitem{french1999catastrophic}
Robert~M French.
\newblock Catastrophic forgetting in connectionist networks.
\newblock {\em Trends in cognitive sciences}, 1999.

\bibitem{gan2023decorate}
Yulu Gan, Yan Bai, Yihang Lou, Xianzheng Ma, Renrui Zhang, Nian Shi, and Lin Luo.
\newblock Decorate the newcomers: Visual domain prompt for continual test time adaptation.
\newblock In {\em AAAI}, 2023.

\bibitem{gao2024consistent}
Zhanxin Gao, Jun Cen, and Xiaobin Chang.
\newblock Consistent prompting for rehearsal-free continual learning.
\newblock In {\em CVPR}, 2024.

\bibitem{gopalakrishnan2022knowledge}
Saisubramaniam Gopalakrishnan, Pranshu~Ranjan Singh, Haytham Fayek, Savitha Ramasamy, and Arulmurugan Ambikapathi.
\newblock Knowledge capture and replay for continual learning.
\newblock In {\em WACV}, 2022.

\bibitem{he2022masked}
Kaiming He, Xinlei Chen, Saining Xie, Yanghao Li, Piotr Doll{\'a}r, and Ross Girshick.
\newblock Masked autoencoders are scalable vision learners.
\newblock In {\em CVPR}, 2022.

\bibitem{hu2022subdivision}
Shi-Min Hu, Zheng-Ning Liu, Meng-Hao Guo, Jun-Xiong Cai, Jiahui Huang, Tai-Jiang Mu, and Ralph~R Martin.
\newblock Subdivision-based mesh convolution networks.
\newblock {\em TOG}, 2022.

\bibitem{kingma2014adam}
Diederik~P Kingma.
\newblock Adam: A method for stochastic optimization.
\newblock {\em ICLR}, 2015.

\bibitem{kirkpatrick2017overcoming}
James Kirkpatrick, Razvan Pascanu, Neil Rabinowitz, Joel Veness, Guillaume Desjardins, Andrei~A Rusu, Kieran Milan, John Quan, Tiago Ramalho, Agnieszka Grabska-Barwinska, et~al.
\newblock Overcoming catastrophic forgetting in neural networks.
\newblock {\em Proceedings of the national academy of sciences}, 2017.

\bibitem{leestella}
Jaewoo Lee, Jaehong Yoon, Wonjae Kim, Yunji Kim, and Sung~Ju Hwang.
\newblock Stella: Continual audio-video pre-training with spatiotemporal localized alignment.
\newblock In {\em ICML}, 2024.

\bibitem{lee2023pre}
Kuan-Ying Lee, Yuanyi Zhong, and Yu-Xiong Wang.
\newblock Do pre-trained models benefit equally in continual learning?
\newblock In {\em WACV}, 2023.

\bibitem{lesort2020continual}
Timoth{\'e}e Lesort, Vincenzo Lomonaco, Andrei Stoian, Davide Maltoni, David Filliat, and Natalia D{\'\i}az-Rodr{\'\i}guez.
\newblock Continual learning for robotics: Definition, framework, learning strategies, opportunities and challenges.
\newblock {\em Information fusion}, 2020.

\bibitem{li2017learning}
Zhizhong Li and Derek Hoiem.
\newblock Learning without forgetting.
\newblock {\em PAMI}, 2017.

\bibitem{lian2011shape}
Z Lian, A Godil, B Bustos, M Daoudi, J Hermans, S Kawamura, Y Kurita, G Lavoua, P~Dp Suetens, et~al.
\newblock Shape retrieval on non-rigid 3d watertight meshes.
\newblock In {\em Eurographics workshop on 3d object retrieval}, 2011.

\bibitem{liang2022meshmae}
Yaqian Liang, Shanshan Zhao, Baosheng Yu, Jing Zhang, and Fazhi He.
\newblock Meshmae: Masked autoencoders for 3d mesh data analysis.
\newblock In {\em ECCV}, 2022.

\bibitem{liu2021lifelong}
Bo Liu, Xuesu Xiao, and Peter Stone.
\newblock A lifelong learning approach to mobile robot navigation.
\newblock {\em IEEE Robotics and Automation Letters}, 2021.

\bibitem{liu2021l3doc}
Yuyang Liu, Yang Cong, Gan Sun, Tao Zhang, Jiahua Dong, and Hongsen Liu.
\newblock L3doc: Lifelong 3d object classification.
\newblock {\em TIP}, 2021.

\bibitem{liu2023online}
Yaoyao Liu, Yingying Li, Bernt Schiele, and Qianru Sun.
\newblock Online hyperparameter optimization for class-incremental learning.
\newblock In {\em AAAI}, 2023.

\bibitem{liu2021adaptive}
Yaoyao Liu, Bernt Schiele, and Qianru Sun.
\newblock Adaptive aggregation networks for class-incremental learning.
\newblock In {\em CVPR}, 2021.

\bibitem{liu2023tail}
Zuxin Liu, Jesse Zhang, Kavosh Asadi, Yao Liu, Ding Zhao, Shoham Sabach, and Rasool Fakoor.
\newblock Tail: Task-specific adapters for imitation learning with large pretrained models.
\newblock {\em ICLR}, 2024.

\bibitem{masana2022class}
Marc Masana, Xialei Liu, Bart{\l}omiej Twardowski, Mikel Menta, Andrew~D Bagdanov, and Joost Van De~Weijer.
\newblock Class-incremental learning: survey and performance evaluation on image classification.
\newblock {\em PAMI}, 2022.

\bibitem{mcdonnell2024ranpac}
Mark~D McDonnell, Dong Gong, Amin Parvaneh, Ehsan Abbasnejad, and Anton van~den Hengel.
\newblock Ranpac: Random projections and pre-trained models for continual learning.
\newblock {\em NIPS}, 2024.

\bibitem{mirza2022efficient}
M~Jehanzeb Mirza, Marc Masana, Horst Possegger, and Horst Bischof.
\newblock An efficient domain-incremental learning approach to drive in all weather conditions.
\newblock In {\em CVPR}, 2022.

\bibitem{moon2023online}
Jun-Yeong Moon, Keon-Hee Park, Jung~Uk Kim, and Gyeong-Moon Park.
\newblock Online class incremental learning on stochastic blurry task boundary via mask and visual prompt tuning.
\newblock In {\em ICCV}, 2023.

\bibitem{pang2022masked}
Yatian Pang, Wenxiao Wang, Francis~EH Tay, Wei Liu, Yonghong Tian, and Li Yuan.
\newblock Masked autoencoders for point cloud self-supervised learning.
\newblock In {\em ECCV}, 2022.

\bibitem{park2024pre}
Keon-Hee Park, Kyungwoo Song, and Gyeong-Moon Park.
\newblock Pre-trained vision and language transformers are few-shot incremental learners.
\newblock In {\em CVPR}, 2024.

\bibitem{paszke2019pytorch}
Adam Paszke, Sam Gross, Francisco Massa, Adam Lerer, James Bradbury, Gregory Chanan, Trevor Killeen, Zeming Lin, Natalia Gimelshein, Luca Antiga, et~al.
\newblock Pytorch: An imperative style, high-performance deep learning library.
\newblock {\em NIPS}, 2019.

\bibitem{pian2023audio}
Weiguo Pian, Shentong Mo, Yunhui Guo, and Yapeng Tian.
\newblock Audio-visual class-incremental learning.
\newblock In {\em ICCV}, 2023.

\bibitem{prabhu2020gdumb}
Ameya Prabhu, Philip~HS Torr, and Puneet~K Dokania.
\newblock Gdumb: A simple approach that questions our progress in continual learning.
\newblock In {\em ECCV}, 2020.

\bibitem{qi2017pointnet}
Charles~R Qi, Hao Su, Kaichun Mo, and Leonidas~J Guibas.
\newblock Pointnet: Deep learning on point sets for 3d classification and segmentation.
\newblock In {\em CVPR}, 2017.

\bibitem{qi2017pointnet++}
Charles~Ruizhongtai Qi, Li Yi, Hao Su, and Leonidas~J Guibas.
\newblock Pointnet++: Deep hierarchical feature learning on point sets in a metric space.
\newblock {\em NIPS}, 2017.

\bibitem{rebuffi2017icarl}
Sylvestre-Alvise Rebuffi, Alexander Kolesnikov, Georg Sperl, and Christoph~H Lampert.
\newblock icarl: Incremental classifier and representation learning.
\newblock In {\em CVPR}, 2017.

\bibitem{rolnick2019experience}
David Rolnick, Arun Ahuja, Jonathan Schwarz, Timothy Lillicrap, and Gregory Wayne.
\newblock Experience replay for continual learning.
\newblock {\em NIPS}, 2019.

\bibitem{roy2024convolutional}
Anurag Roy, Riddhiman Moulick, Vinay~K Verma, Saptarshi Ghosh, and Abir Das.
\newblock Convolutional prompting meets language models for continual learning.
\newblock In {\em CVPR}, 2024.

\bibitem{szatkowski2024adapt}
Filip Szatkowski, Mateusz Pyla, Marcin Przewi{\k{e}}{\'z}likowski, Sebastian Cygert, Bart{\l}omiej Twardowski, and Tomasz Trzci{\'n}ski.
\newblock Adapt your teacher: Improving knowledge distillation for exemplar-free continual learning.
\newblock In {\em WACV}, 2024.

\bibitem{tan2024cross}
Yuwen Tan and Xiang Xiang.
\newblock Cross-domain few-shot incremental learning for point-cloud recognition.
\newblock In {\em WACV}, 2024.

\bibitem{tran2024text}
Minh-Tuan Tran, Trung Le, Xuan-May Le, Mehrtash Harandi, and Dinh Phung.
\newblock Text-enhanced data-free approach for federated class-incremental learning.
\newblock In {\em CVPR}, 2024.

\bibitem{truong2024conda}
Thanh-Dat Truong, Pierce Helton, Ahmed Moustafa, Jackson~David Cothren, and Khoa Luu.
\newblock Conda: Continual unsupervised domain adaptation learning in visual perception for self-driving cars.
\newblock In {\em CVPR}, 2024.

\bibitem{wang2022foster}
Fu-Yun Wang, Da-Wei Zhou, Han-Jia Ye, and De-Chuan Zhan.
\newblock Foster: Feature boosting and compression for class-incremental learning.
\newblock In {\em ECCV}, 2022.

\bibitem{wang2024comprehensive}
Liyuan Wang, Xingxing Zhang, Hang Su, and Jun Zhu.
\newblock A comprehensive survey of continual learning: theory, method and application.
\newblock {\em PAMI}, 2024.

\bibitem{wang2019dynamic}
Yue Wang, Yongbin Sun, Ziwei Liu, Sanjay~E Sarma, Michael~M Bronstein, and Justin~M Solomon.
\newblock Dynamic graph cnn for learning on point clouds.
\newblock {\em TOG}, 2019.

\bibitem{wang2022dualprompt}
Zifeng Wang, Zizhao Zhang, Sayna Ebrahimi, Ruoxi Sun, Han Zhang, Chen-Yu Lee, Xiaoqi Ren, Guolong Su, Vincent Perot, Jennifer Dy, et~al.
\newblock Dualprompt: Complementary prompting for rehearsal-free continual learning.
\newblock In {\em ECCV}, 2022.

\bibitem{wang2022learning}
Zifeng Wang, Zizhao Zhang, Chen-Yu Lee, Han Zhang, Ruoxi Sun, Xiaoqi Ren, Guolong Su, Vincent Perot, Jennifer Dy, and Tomas Pfister.
\newblock Learning to prompt for continual learning.
\newblock In {\em CVPR}, 2022.

\bibitem{wu20153d}
Zhirong Wu, Shuran Song, Aditya Khosla, Fisher Yu, Linguang Zhang, Xiaoou Tang, and Jianxiong Xiao.
\newblock 3d shapenets: A deep representation for volumetric shapes.
\newblock In {\em CVPR}, 2015.

\bibitem{yang2023geometry}
Yuwei Yang, Munawar Hayat, Zhao Jin, Chao Ren, and Yinjie Lei.
\newblock Geometry and uncertainty-aware 3d point cloud class-incremental semantic segmentation.
\newblock In {\em CVPR}, 2023.

\bibitem{ye2024continual}
Yiwen Ye, Yutong Xie, Jianpeng Zhang, Ziyang Chen, Qi Wu, and Yong Xia.
\newblock Continual self-supervised learning: Towards universal multi-modal medical data representation learning.
\newblock In {\em CVPR}, 2024.

\bibitem{yuan2023peer}
Liangqi Yuan, Yunsheng Ma, Lu Su, and Ziran Wang.
\newblock Peer-to-peer federated continual learning for naturalistic driving action recognition.
\newblock In {\em CVPR}, 2023.

\bibitem{yue2024mmal}
Xianghu Yue, Xueyi Zhang, Yiming Chen, Chengwei Zhang, Mingrui Lao, Huiping Zhuang, Xinyuan Qian, and Haizhou Li.
\newblock Mmal: Multi-modal analytic learning for exemplar-free audio-visual class incremental tasks.
\newblock In {\em ACMMM}, 2024.

\bibitem{zamorski2023continual}
Maciej Zamorski, Micha{\l} Stypu{\l}kowski, Konrad Karanowski, Tomasz Trzci{\'n}ski, and Maciej Zi{\k{e}}ba.
\newblock Continual learning on 3d point clouds with random compressed rehearsal.
\newblock {\em Computer Vision and Image Understanding}, 2023.

\bibitem{zhang2022pointclip}
Renrui Zhang, Ziyu Guo, Wei Zhang, Kunchang Li, Xupeng Miao, Bin Cui, Yu Qiao, Peng Gao, and Hongsheng Li.
\newblock Pointclip: Point cloud understanding by clip.
\newblock In {\em CVPR}, 2022.

\bibitem{zhou2024revisiting}
Da-Wei Zhou, Zi-Wen Cai, Han-Jia Ye, De-Chuan Zhan, and Ziwei Liu.
\newblock Revisiting class-incremental learning with pre-trained models: Generalizability and adaptivity are all you need.
\newblock {\em IJCV}, 2024.

\bibitem{zhou2024continual}
Da-Wei Zhou, Hai-Long Sun, Jingyi Ning, Han-Jia Ye, and De-Chuan Zhan.
\newblock Continual learning with pre-trained models: A survey.
\newblock In {\em IJCAI}, 2024.

\bibitem{zhou2024class}
Da-Wei Zhou, Qi-Wei Wang, Zhi-Hong Qi, Han-Jia Ye, De-Chuan Zhan, and Ziwei Liu.
\newblock Class-incremental learning: A survey.
\newblock {\em PAMI}, 2024.

\bibitem{zhuang2023gkeal}
Huiping Zhuang, Zhenyu Weng, Run He, Zhiping Lin, and Ziqian Zeng.
\newblock Gkeal: Gaussian kernel embedded analytic learning for few-shot class incremental task.
\newblock In {\em CVPR}, 2023.

\bibitem{zhuang2022acil}
Huiping Zhuang, Zhenyu Weng, Hongxin Wei, Renchunzi Xie, Kar-Ann Toh, and Zhiping Lin.
\newblock Acil: Analytic class-incremental learning with absolute memorization and privacy protection.
\newblock {\em NIPS}, 2022.

\end{thebibliography}
}

\end{document}


\maketitle
\thispagestyle{empty}
\pagestyle{empty} 

\section{Proof of Theorem 1}

\noindent In phase $n-1$, we have the weight matrix as follows:
\begin{equation*}
\hat{\boldsymbol{W}}_{n-1} = \left(\sum_{n'=0}^{n-1} \boldsymbol{A}_{n'} + \eta \boldsymbol{I}\right)^{-1} \left(\sum_{n'=0}^{n-1} \boldsymbol{C}_{n'}\right) \tag{1}
\end{equation*}
\vspace{4pt}

\noindent where \(\boldsymbol{A}_{n'}\) and \(\boldsymbol{C}_{n'}\) represent the auto-correlation and cross-correlation matrices. These matrices are defined as:
\begin{equation*}
\boldsymbol{A}_{n'} = \left(\boldsymbol{F}_{n'}^{\mathrm{RP}}\right)^{\top} \boldsymbol{F}_{n'}^{\mathrm{RP}}, \quad
\boldsymbol{C}_{n'} = \left(\boldsymbol{F}_{n'}^{\mathrm{RP}}\right)^{\top} \boldsymbol{Y}_{n'}^{\mathrm{train}} \tag{2}
\end{equation*}
\vspace{4pt}

\noindent Similarly, at phase $n$, we have:
\begin{equation*}
\hat{\boldsymbol{W}}_n = \left(\sum_{n'=0}^{n} \boldsymbol{A}_{n'} + \eta \boldsymbol{I}\right)^{-1} \left(\sum_{n'=0}^{n} \boldsymbol{C}_{n'}\right) \tag{3}
\end{equation*}
\vspace{4pt}

\noindent In the paper, we define the regularized auto-correlation matrix at phase \(n-1\) as:
\begin{equation*}
\boldsymbol{R}_{n-1} = \left(\sum_{n'=0}^{n-1} \boldsymbol{A}_{n'} + \eta \boldsymbol{I}\right)^{-1} \tag{4}
\end{equation*}
\vspace{4pt}

\noindent At phase $n$, we have:
\begin{align*}
\boldsymbol{R}_n &= \left( \sum_{n^{\prime}=0}^n \boldsymbol{A}_{n^{\prime}} + \eta \boldsymbol{I} \right)^{-1}\\
&= \left( \sum_{n^{\prime}=0}^{n-1} \boldsymbol{A}_{n^{\prime}} + \boldsymbol{A}_n + \eta \boldsymbol{I} \right)^{-1} 
= \left( \boldsymbol{R}_{n-1}^{-1} + \boldsymbol{A}_n \right)^{-1} 
= \left( \boldsymbol{R}_{n-1}^{-1} + \left( \boldsymbol{F}_n^{\mathrm{RP}} \right)^{\top} \boldsymbol{F}_n^{\mathrm{RP}} \right)^{-1} \tag{5}
\end{align*}
\vspace{4pt}

\noindent According to the Woodbury matrix identity, we have:
\begin{equation*}
(\boldsymbol{A} + \boldsymbol{U} \boldsymbol{C} \boldsymbol{V})^{-1} = \boldsymbol{A}^{-1} - \boldsymbol{A}^{-1} \boldsymbol{U} \left(\boldsymbol{V} \boldsymbol{A}^{-1} \boldsymbol{U} + \boldsymbol{C}^{-1}\right) \boldsymbol{V} \boldsymbol{A}^{-1} \tag{6}
\end{equation*}
\vspace{4pt}

\noindent By setting \( \boldsymbol{A} = \boldsymbol{R}_{n-1}^{-1} \), \( \boldsymbol{U} = \left(\boldsymbol{F}_n^{\mathrm{RP}}\right)^\top \), \( \boldsymbol{C} = \boldsymbol{I} \), and \( \boldsymbol{V} = \boldsymbol{F}_n^{\mathrm{RP}} \) in Equation (5), this leads to the following update:
\begin{equation*}
\boldsymbol{R}_n = \boldsymbol{R}_{n-1} - \boldsymbol{R}_{n-1} \left(\boldsymbol{F}_n^{\mathrm{RP}}\right)^\top
\left(\boldsymbol{F}_n^{\mathrm{RP}} \boldsymbol{R}_{n-1} \left(\boldsymbol{F}_n^{\mathrm{RP}}\right)^\top + \boldsymbol{I}\right)^{-1} \boldsymbol{F}_n^{\mathrm{RP}} \boldsymbol{R}_{n-1} \tag{7}
\end{equation*}
\vspace{4pt}

\noindent This expression exemplifies the recursive nature of the updates for \(\boldsymbol{R}_n\), using the prior matrix \(\boldsymbol{R}_{n-1}\) and incorporating new feature data \(\boldsymbol{F}_n^{\mathrm{RP}}\). This proof completes the mathematical formulation for the recursive calculation of \(\boldsymbol{R}_n\).\vspace{1\baselineskip}

\noindent Next, we derive the recursive calculation of \( \hat{\boldsymbol{W}}_n \). To this end, we first recursively calculate \( \sum_{n'=0}^{n} \boldsymbol{C}_{n'} \), i.e.,
\begin{equation*}
\sum_{n'=0}^{n} \boldsymbol{C}_{n'} = \sum_{n'=0}^{n-1} \boldsymbol{C}_{n'} + \boldsymbol{C}_n = \sum_{n'=0}^{n-1} \boldsymbol{C}_{n'} + \left(\boldsymbol{F}_n^{\mathrm{RP}}\right)^\top \boldsymbol{Y}_n^{\mathrm{train}} \tag{8}
\end{equation*}
\vspace{4pt}

\noindent Let \( \boldsymbol{K}_n = \left(\boldsymbol{F}_n^{\mathrm{RP}} \boldsymbol{R}_{n-1} \left(\boldsymbol{F}_n^{\mathrm{RP}}\right)^\top + \boldsymbol{I}\right)^{-1} \). Since
\begin{equation*}
\boldsymbol{I} = \boldsymbol{K}_n \boldsymbol{K}_n^{-1} = \boldsymbol{K}_n \left(\boldsymbol{F}_n^{\mathrm{RP}} \boldsymbol{R}_{n-1} \left(\boldsymbol{F}_n^{\mathrm{RP}}\right)^\top + \boldsymbol{I}\right) \tag{9}
\end{equation*}
\vspace{4pt}

\noindent we have
\begin{equation*}
\boldsymbol{K}_n = \boldsymbol{I} - \boldsymbol{K}_n \boldsymbol{F}_n^{\mathrm{RP}} \boldsymbol{R}_{n-1} \left(\boldsymbol{F}_n^{\mathrm{RP}}\right)^\top.
\end{equation*}
\vspace{5pt}

\noindent Therefore,
\begin{equation*}
\begin{aligned}
& \boldsymbol{R}_{n-1} \left(\boldsymbol{F}_n^{\mathrm{RP}}\right)^\top \left(\boldsymbol{F}_n^{\mathrm{RP}} \boldsymbol{R}_{n-1} \left(\boldsymbol{F}_n^{\mathrm{RP}}\right)^\top + \boldsymbol{I}\right)^{-1} \\
&= \boldsymbol{R}_{n-1} \left(\boldsymbol{F}_n^{\mathrm{RP}}\right)^\top \boldsymbol{K}_n \\
&= \boldsymbol{R}_{n-1} \left(\boldsymbol{F}_n^{\mathrm{RP}}\right)^\top \left(\boldsymbol{I} - \boldsymbol{K}_n \boldsymbol{F}_n^{\mathrm{RP}} \boldsymbol{R}_{n-1} \left(\boldsymbol{F}_n^{\mathrm{RP}}\right)^\top \right) \\
&= \left(\boldsymbol{R}_{n-1} - \boldsymbol{R}_{n-1} \left(\boldsymbol{F}_n^{\mathrm{RP}}\right)^\top \boldsymbol{K}_n \boldsymbol{F}_n^{\mathrm{RP}} \boldsymbol{R}_{n-1}\right)\left(\boldsymbol{F}_n^{\mathrm{RP}}\right)^\top \\
&= \boldsymbol{R}_n \left(\boldsymbol{F}_n^{\mathrm{RP}}\right)^\top
\end{aligned}
\tag{10}
\end{equation*}
\vspace{5pt}

\noindent Consequently, the updated weight matrix \(\hat{\boldsymbol{W}}_n\) can be expressed as follows:
\begin{equation*}
\begin{aligned}
\hat{\boldsymbol{W}}_n & = \boldsymbol{R}_n \sum_{n'=0}^{n} \boldsymbol{C}_{n'} \\
& = \boldsymbol{R}_n \left(\sum_{n'=0}^{n-1} \boldsymbol{C}_{n'} + \left(\boldsymbol{F}_n^{\mathrm{RP}}\right)^\top \boldsymbol{Y}_n^{\mathrm{train}}\right) \\
& = \boldsymbol{R}_n \sum_{n'=0}^{n-1} \boldsymbol{C}_{n'} + \boldsymbol{R}_n \left(\boldsymbol{F}_n^{\mathrm{RP}}\right)^\top \boldsymbol{Y}_n^{\mathrm{train}}
\end{aligned}
\tag{11}
\end{equation*}
\vspace{5pt}

\noindent By substituting (7) into \( \boldsymbol{R}_n \sum_{n'=0}^{n-1} \boldsymbol{C}_{n'} \), we have:
\begin{equation*}
\begin{aligned}
&\boldsymbol{R}_n \sum_{n'=0}^{n-1} \boldsymbol{C}_{n'} \\
&= \boldsymbol{R}_{n-1} \sum_{n'=0}^{n-1} \boldsymbol{C}_{n'}  - \boldsymbol{R}_{n-1} \left(\boldsymbol{F}_n^{\mathrm{RP}}\right)^\top 
\left(\boldsymbol{F}_n^{\mathrm{RP}} \boldsymbol{R}_{n-1} \left(\boldsymbol{F}_n^{\mathrm{RP}}\right)^\top + \boldsymbol{I}\right)^{-1} \boldsymbol{F}_n^{\mathrm{RP}} \boldsymbol{R}_{n-1} \sum_{n'=0}^{n-1} \boldsymbol{C}_{n'} \\
& = \hat{\boldsymbol{W}}_{n-1} - \boldsymbol{R}_{n-1} \left(\boldsymbol{F}_n^{\mathrm{RP}}\right)^\top 
\left(\boldsymbol{F}_n^{\mathrm{RP}} \boldsymbol{R}_{n-1} \left(\boldsymbol{F}_n^{\mathrm{RP}}\right)^\top + \boldsymbol{I}\right)^{-1} \boldsymbol{F}_n^{\mathrm{RP}} \hat{\boldsymbol{W}}_{n-1}
\end{aligned}
\tag{12}
\end{equation*}
\vspace{5pt}

\noindent According to (10), (12) can be rewritten as:
\begin{equation*}
\boldsymbol{R}_n \sum_{n'=0}^{n-1} \boldsymbol{C}_{n'} = \hat{\boldsymbol{W}}_{n-1} - \boldsymbol{R}_n \left(\boldsymbol{F}_n^{\mathrm{RP}}\right)^\top \boldsymbol{F}_n^{\mathrm{RP}} \hat{\boldsymbol{W}}_{n-1} \tag{13}
\end{equation*}
\vspace{4pt}

\noindent By inserting (13) into (11), we have:
\begin{equation*}
\begin{aligned}
\hat{\boldsymbol{W}}_n &= \hat{\boldsymbol{W}}_{n-1} - \boldsymbol{R}_n \left(\boldsymbol{F}_n^{\mathrm{RP}}\right)^\top \boldsymbol{F}_n^{\mathrm{RP}} \hat{\boldsymbol{W}}_{n-1} + \boldsymbol{R}_n \left(\boldsymbol{F}_n^{\mathrm{RP}}\right)^\top \boldsymbol{Y}_n^{\mathrm{train}} \\
&= \hat{\boldsymbol{W}}_{n-1} - \boldsymbol{R}_n \boldsymbol{A}_n \hat{\boldsymbol{W}}_{n-1} + \boldsymbol{R}_n \boldsymbol{C}_n
\end{aligned}
\tag{14}
\end{equation*}

\vspace{4pt}

\noindent which proves the recursive calculation of \( \hat{\boldsymbol{W}}_n \).